\RequirePackage{amsmath}
\documentclass{llncs}

\pdfoutput=1

\usepackage[english]{babel}
\usepackage[utf8]{inputenc}
\usepackage{xcolor}
\usepackage{amsmath}
\usepackage{amssymb}
\usepackage{graphicx}
\usepackage[colorinlistoftodos]{todonotes}
\usepackage{url}
\usepackage{amsfonts}
\usepackage{paralist}
\usepackage{listings}
\usepackage{courier}
\usepackage{caption}
\usepackage{subcaption}
\usepackage{booktabs} %
\usepackage[T1]{fontenc}
\usepackage[scaled=0.9]{beramono}
\usepackage[tracking=true]{microtype}

\newcommand\Small{\fontsize{8}{10}\selectfont}
\newcommand*\LSTfont{%
  \Small\ttfamily\SetTracking{encoding=*}{-50}\lsstyle}

\usepackage{hyperref}
\usepackage{cleveref}

\crefname{theorem}{theorem}{theorems}
\Crefname{theorem}{Theorem}{Theorems}

\crefname{definition}{definition}{definitions}
\Crefname{definition}{Definition}{Definitions}

\newcommand{\emptyProtoID}[0]{\mathtt{P}_{\emptyset}}

\begin{document}

\title{Knowledge Representation on the Web revisited: \newline 
       Tools for Prototype Based Ontologies}
\titlerunning{KR on the Web revisited: Tools for Prototypes}  %

\author{Michael Cochez\inst{1,2,4} \and Stefan Decker\inst{1,2} \and Eric Prud'hommeaux\inst{3}}
\authorrunning{M.~Cochez, S.~Decker, E.~Prud'hommeaux} %

\institute{
Fraunhofer Institute for Applied Information Technology FIT\\
DE-53754 Sankt Augustin, Germany\\
\email{\{stefan.decker,michael.cochez\}@fit.fraunhofer.de}
\and
RWTH Aachen University, Informatik 5\\
DE-52056 Aachen, Germany\\
\and
World Wide Web Consortium (W3C)\\
Stata Center, MIT\\
\email{eric@w3.org}
\and
University of Jyvaskyla, Department of Mathematical Information Technology\\
FI-40014 University of Jyväskylä, Finland\\
}

\maketitle              %

\begin{abstract}
In recent years RDF and OWL have become the most common knowledge representation languages in use on the Web, propelled by the recommendation of the W3C. 
In this paper we present a practical implementation of a different kind of knowledge representation based on Prototypes. 
In detail, we present a concrete syntax easily and effectively parsable by applications.   
We also present extensible implementations of a prototype knowledge base, specifically designed for storage of Prototypes.
These implementations are written in Java and can be extended by using the implementation as a library.
Alternatively, the software can be deployed as such.
Further, results of benchmarks for both local and web deployment are presented.
This paper augments a research paper, in which we describe the more theoretical aspects of our Prototype system.
\end{abstract}

\section{Introduction}
Recently, we proposed Prototypes as a way to represent knowledge on the web~\cite{researchPaper}\footnote{
Please use that paper as a reference for all definitions.}.
That paper has its focus on theoretical aspects and analysis.
In this resource paper we describe the tools we developed to deploy prototypes.
First, in \cref{sec:standalone} we present the implementation of a knowledge base, based on the implementation of a Java interface, which can be used for storing prototypes.
Then, we reuse this system to show how the knowledge base can be used in remote and distributed settings (\cref{sec:distributed}).
Each of the sections includes some amount of benchmarking to give the reader an impression of the practical re-usability of the provided solutions.
We do assume that the reader has some familiarity with the ideas behind prototypes.
The implementation and the code used for the benchmarks is licensed under the LGPLv3 license and can be downloaded from \url{https://github.com/miselico/knowledgebase}.

\section{A Standalone Knowledge Base}
\label{sec:standalone}

A prototype knowledge base (KB) consists of a collection of prototypes.
To mirror this, the IKnowledge interface, which we define as the basis for a KB, has only one method which must\footnote{Other methods are Java 8 default methods.} be implemented. The signature of this method is \lstinline$Optional<? extends Prototype> isDefined(ID id);$.
The provided Java source code contains five implementations of this interface, namely 
\begin{description}
\item[EmptyKnowledgeBase] is a KB without any content. However, as per the definition, it still contains the empty prototype $\emptyProtoID$.
\item[PredefinedKB] is a KB containing string and integer constants and is described in more details in this section.
\item[KnowledgeBase] stores prototypes. It can be constructed using another \lstinline$IKnowledgeBase$ as a basis. The underlying basis will be queried in case the requested prototype is not directly defined in the \lstinline$KnowledgeBase$.
\item[RemoteKB] gets its prototypes from a remote KB. This implementation is described further in~\cref{sec:remote}.
\item[ChainedKB] is an \lstinline$IknowledgeBase$ which connects multiple \lstinline$IKnowledgeBase$s together. 
The KBs are checked in turn until one where the Prototype is defined is found. If none is found, an empty \lstinline$Optional$ is returned, indicating that no Prototype could be found.
\end{description}

Each prototype consists of four components, namely 
 \begin{inparaenum}[\itshape 1\upshape)]
\item its own ID, 
\item the ID of its base, 
\item the change set for adding parts, and
\item the change set for removing parts~\cite{researchPaper}.
\end{inparaenum}
This structure is closely mimicked in our implementation. The IDs are essentially represented using String types and the change sets using multimaps (i.e., maps which can associate multiple values for a given key).
The formal definition allows the creation of a prototype which would remove all values for a given property.
According to the theoretical definition, that computation would involve the set of all possible IDs by enumeration, which is unfeasible. 
Hence %
our implementation has two distinct changeset implementations and treats the `remove all' as a special case which does not require enumeration.

Another aspect which a concrete implementation should cover is the use of literals.
At its current state the formal Prototype KB definition does not support literals as the value of a property.
Instead, one has to represent a literal by using an agreed prototype.
Therefore, we designed \lstinline$PredefinedKB$ which acts like a KB which implicitly contains all possible string and integer literals encoded as prototypes.
The extension of the supported literal types to any type is facilitated.
\subsection{Consistency Checking}

When a KB is created out of a set of prototypes, it should be checked 
that the result is in accordance with our Prototype Knowledge Base definition 
\cite[Definition 4]{researchPaper}.
In this paper we say that the KB must be checked for consistency. 
This consistency check is performed in the \lstinline$KnowledgeBase$ implementation.
First, all IDs and property names must be valid absolute IRIs, which is enforced by the type system.
Next, if there is any prototype with a definition involving an $ID$ which cannot be found in the KB, then the creation will be refused.
Then, it is checked whether all inheritance chains eventually (recursively) end up at the empty prototype.
If also that is the case then a check is performed to ensure that no ID is used twice (this includes checking the underlying KB).
In practice, one might want to remove this last check and the issue of duplicates could be resolved in favor of prototypes in a given KB.
This is possible using the \lstinline$ChainedKB$.

The design of our software helps to build KB which are consistent.
The KB provides a builder class which can be used for construction.
Further, the KB itself is completely immutable.
Changes are made by creating a new KB.
This ensures the consistency at any point in time.

\subsection{Fixpoint Computation}

Given a knowledge base $KB_{o}$ we implemented a method to compute its interpretation $KB_{n}$. 
This interpretation contains for each prototype definition with ID $id$ in $KB_o$ a new prototype definition of the form $(id, (\text{ \url{PROTO:P_0} }, ADD, \emptyset))$.
Where $ADD$ is such that under an interpretation $I_{KB}(KB_o) = I_{KB} (KB_n)$.
This boils down to computing the fixpoint for each of the prototype expressions.
However, a direct implementation of the definition would not work since there is a universal quantification over all IDs (an infinite set).
Hence, we implement this such that simple change expressions are created only for these IDs which are actually used.

We implemented both the consistency check and the fixpoint computation in a scalable fashion.
For both the consistency check and the computation of the fixpoints, the implementation is optimized such that it will not compute things twice.
When the fixpoint for a prototype $P_1$ has already been computed, then the computation of the fixpoint for a prototype $P_2$ with $base$ $P_1$ will reuse this result.
Similarly during the consistency check for recursive derivation from $\emptyProtoID$: if it is already known that a prototype $P_1$ derives recursively from $\emptyProtoID$, then we reuse this information to conclude that $P_2$ with base $P_1$ derives from $\emptyProtoID$.
Next, we will introduce the data sets and the benchmarks in which they are used.

\subsection{Data Sets and Benchmarks}
\label{sec:datasets}

Since the prototype system is new there are no existing real-world dataset which make use of its features.
It would be possible to use existing RDF datasets as prototypes, but it would result in a KB which does not use the inheritance feature specific to prototypes.
Therefore, we decided to use synthetic data sets for our benchmarks.
We created three types of data sets in three different sizes, resulting in nine data sets altogether.
Note that we do not use the remove capabilities in our data sets. 
This is not required since a remove will only reduce the burden on the system.
An overview of the datasets can be found in \cref{tab:datasets}.

\begin{table*}[t]
\caption{An overview of the data sets. The numbers between brackets indicate the different size parameters used to generate the data sets (see \cref{sec:datasets} for more information).
The table shows the amount of prototypes and the average number (and st. dev.) of properties in the add set of the prototypes.
Below we will refer to the data sets with their initial letters only.
}
\label{tab:datasets} 
\centering
\setlength{\tabcolsep}{4pt}
\makebox[\textwidth][c]{ %
\begin{tabular}{lcccccc}
\toprule
Data set & \multicolumn{3}{c}{prototypes} & \multicolumn{3}{c}{properties per prototype}   \\ 
\cmidrule(r){2-4}  \cmidrule(r){5-7}
\textbf{ba}seline (19/20/21) &1,048,575 & 2,097,151 & 4,194,303 & 0 & 0 & 0 \\
\textbf{bl}ocks (10/20/30)     &1,000,000 & 2,000,000 & 3,000,000 & 1 & 1 & 1\\
\textbf{inc}remental (1/2/3) & 1,000,000 & 2,000,000 & 3,000,000 & 2.0 $\pm$ 1.4 &2.0 $\pm$ 1.4 &2.0 $\pm$ 1.4 \\
\bottomrule 
\end{tabular} 
}
\end{table*}

The first type of data sets does have beneficial properties for consistency checking and fixpoint computation.
Further it does not have any properties attached to the proptotypes.
Hence, we will call these \emph{baseline} data sets.
To generate the data we start with one prototype which derives from $\emptyProtoID$, next we create two prototypes which derive form the one, then we create four prototypes which derive from these two, and so on until we create $2^{n}$ prototypes which derive from $2^{n-1}$ (for $n \in \{19,20,21\}$). 

For the second type we change the set-up to be less ideal and introduce properties.
We create $10$, $20$, and $30$ blocks of $100,000$ prototypes and hence this type will be called \emph{blocks}.
All prototypes in each block derive from a randomly chosen prototype in a lower block.
Then, each of the prototypes has a property with a value randomly chosen from the block below. 
In the lowest block, the base is always $\emptyProtoID$ and the value for the property is always the same fixed prototype.

The third type of data sets, which we call \emph{incremental}, is more demanding for the fixpoint computation and consistency check.
This time we add 1, 2, and 3 million prototypes to the KB, one at a time.
Each prototype gets a randomly selected earlier created one as its base.
Furthermore, each prototype gets between $0$ and $4$ properties chosen from $10$ distinct ones (with replacement).
The value of each property is chosen randomly among the prototypes.

\vspace{-1em}
\subsubsection{Results:}

For each data set we measure how long it takes to perform the consistency check and to compute the fixpoint of all prototypes (i.e., compute the whole interpretation).
We also measure the final average number of properties per prototype.
These results can be found in \cref{tab:experiments}.

\begin{table*}[t]
\caption{The outcomes of the benchmark. 
For each dataset the table shows how long the consistency check took to complete and the time needed for the computation of the fixpoint.
The last three columns show the average number (and standard deviation) of properties after computation of the fixpoint.}
\label{tab:experiments} 
\centering
\setlength{\tabcolsep}{4pt}
\makebox[\textwidth][c]{ %
\begin{tabular}{lccccccccc}
\toprule
Data set & \multicolumn{3}{c}{consistency(ms)} & \multicolumn{3}{c}{fixpoint(ms)} &\multicolumn{3}{c}{prop. per prototype in fp.}   \\ 
\cmidrule(r){2-4}  \cmidrule(r){5-7} \cmidrule(r){8-10}
\textbf{ba} (19/20/21) & 2,659 & 4,083 & 8,150 & 5,281  & 7,344 & 15,055 & 0 & 0 & 0\\
\textbf{bl} (10/20/30) & 3,517 & 6,195 & 9,278 & 12,740 & 27,367 & 50,003 & 5.5 $\pm$ 2.9 & 10.5 $\pm$ 5.8 & 15.5 $\pm$ 8.6\\
\textbf{inc} (1/2/3)   &  4,580 &8,469 & 10,436 & 23,597 &57,151 &94,702 & 26.7 $\pm$ 9.0& 27.3 $\pm$ 9.1& 30.0 $\pm$ 9.6\\
\bottomrule 
\end{tabular} 
}
\end{table*}

As can be seen from the table, the consistency check scales linear with the number of prototypes in the system.
In some cases it seems like the behavior is even sub-linear. This is likely caused by just-in-time compilation.

For the fixpoint, the baseline dataset provides close to linear performance, which is expected.
Again, larger sets seem to compute even faster (or rather, the small set is handled slower because the JIT compiler has not yet optimized the code).
The blocks and incremental experiments also show the expected scalability. 
They do, however, not have a linear scaling because in the larger experiments the numer of properties per prototype in the fixpoints is larger.

The results are obtained by running the code on a single `Intel(R) Xeon(R) E5-2670 @ 2.60GHz' core (using \emph{taskset}).
To keep results comparable we allowed the JVM to use large amounts of memory.
The memory was mainly used to store the fixpoint of the KB, something one would in practice rarely keep in memory.
These results show that the prototype system can scale well, even to millions of prototypes.

\section{Distributed Knowledge Bases}
\label{sec:remote}
\label{sec:distributed}
Our goal is to create a KB which can be used on the web.
Hence, it is not sufficient to show that our implementation can be used locally.
In this section we present how we implemented the client-server version of our KB.
In our implementation we use the ubiquitous HTTP protocol and well known data formats. 
We also illustrate that KBs can have explicit links to each other trough the \texttt{Link} header of the HTTP protocol.
To show the knowledge sharing scenario in action we perform benchmarks in which we query the KB in a simulated web environment.

\subsection{Prototype Serialization and Joining}
To communicate the Prototypes between server and client we want to use a language which is platform independent.
Furthermore, the serialization should be reasonably easy to parse using different modern programming languages.
Despite the simple textual serialization already available in the software and demonstrated in the research paper~\cite{researchPaper}, we choose to implement JSON serialization for the client--server interaction.
This would also enable more straightforward integration with Javascript clients (and servers) at a later point.
The JSON serialization itself is straightforward. A prototype is converted to the following structure:
\begin{lstlisting}
{"id":"theID", "base":"baseID", "add":{"propA":["id1", ...], ...}, 
"rem":{"propB":["id3", ...], ...}, "remAll":{"propC", ...}}
\end{lstlisting}

\subsubsection{Joining of Prototype Definitions} is needed when retrieving the representation of a prototype from multiple sources.
In general the approach to this problem is dependent on the data sources and the amount of trust the client has in them.
Imagine that one queries data source A for information about Germany.
One of the properties of Germany returned by A is that the country has 80M inhabitants.
When service B is asked about the same prototype a population of 20M is claimed.
Now, the client has a specification of the schema it wants the countries to fulfill.
Concrete, it could be that a country can only have one number for the population count property.
Hence, the application would choose for the amount from the most trusted source.
Several implementations of joining the changesets of prototypes are provided.

\subsection{Deployment On Web Architecture}
To serve prototypes on the web we use the HTTP protocol.
To get the (serialized form of) the prototype, one needs to send a GET request with the protoype ID as a query parameter with the name $p$.
For example, if the server is located at \url{http://example.com/} and one wants to request the prototype \url{isbn:123-4-56-789012-3}, then the request URL will be \url{http://example.com?p=isbn%3A123-4-56-789012-3}. 
We also implemented a way to serve fixpoints of prototypes.
Since we are using HTTP, we can also use the existing optimizations and caching startegies available.
From the server perspective, we use gzip compression in case the client supports it.
Further, the server indicates how long the prototype will remain unchanged using the Cache-Control header \cite{rfc7234}.
Besides, the ETag header \cite{rfc7232} is used; if the client wants to use the prototype but the cache time has expired, then it only needs to check with the server whether the ETag has changed to know whether it is up-to-date.
The server implementation uses an embedded Jetty server which can be configured as desired. For instance, it is possible to deploy the implemented handler using HTTPS or HTTP/2. 
The client side (\lstinline|RemoteKB|) supports content compression, implements the caching of prototypes, and uses the ETag information. 
Besides, HTTP connections are reused for multiple prototypes in order to avoid the TCP handshake.
Further, the client (and server) can handle multiple requests concurrently.

The \texttt{Link: rel="alternate"} HTTP header~\cite{rfc5988} is used to indicate that other providers might have more information about a given prototype.
Concrete, if a client asks for a prototype from service provider A, A might tell that B might have a definition of the given prototype as well.
An application can then, in turn, get the definition from service B. 
We refrain from defining the exact meaning of the link between the two prototypes and rely on the application to choose its interpretation.
We could have chosen to use the semantics of \url{owl:sameAs}, but this has lead to confusion and misuse in the past (see also~\cite{halpin2010owl}).

The server also supports the retrieval of multiple prototypes at once.
However, in that mode ETags and alternates will not work.
Furthermore, if any of the prototypes requested is not found, the server will return an error code.
Having this functionality is reasonable since the payload of the responses is relatively small and hence the serving time is dominated by the round trip time.

\subsection{Benchmarks}
In order to get repeatable results in our benchmarks, we work in a simulated web environment.
For the simulation of a web environment there are two essential components.
First, there are delays and losses on the network and second limited transmission rates.
To simulate the delays and losses we use the netem\footnote{\url{http://www.linuxfoundation.org/collaborate/workgroups/networking/netem}} kernel component available in the Linux kernel.
We attempt to obtain realistic settings by performing prior experiments in which we measure these factors when connecting from Finland (Europe) to New York city (USA).
This way we found an average round trip time of 184.7 ms (min: 184.4 ms, max: 185.1 ms, stdev: 0.3 ms).
We did not get any packet loss during out experiments, but will anyway use a package loss of 0.04\% (i.e 4 in 10,000 packets got lost in transmission).
We arbitrary capped the bandwidth to 1024 KBit/s using token bucket filtering. This speed is arguably on the lower side, but shows that the KB can also be used when only limited bandwidth is available.

In this benchmark we use the blocks(30) data set as presented in \cref{sec:datasets} and perform benchmarks in two environments.
In the first and second benchmark, the network speed is virtually unconstrained (both server and client are on the same host).
The difference between these two experiments is that first we only allow one concurrent request, while in the second one we allow up to 100.
In the third experiment client and server are both deployed on their own virtual machine with the network modifications described above, the client can make up to 100 requests concurrently.

In the benchmarks the client selects random prototypes and requests the fixpoint from the server.
The timing for different amounts of prototypes can be found from~\cref{tab:remote}.
As can be seen, it is beneficial to perform multiple requests simultaneously.
Further, the network delay has a major impact on the timings.
It becomes clear that requesting a couple of thousand fixpoints from the server does not really put it under stress.
This can be seen from the second benchmark where hundred requests are send at the same time: the time to get 10,000 fixpoints is only half a second longer than to get 1,000 fixpoints.
For the remote case we further observe a linear relation between the time needed and number of prototypes and note that in our simulated web environment we are able to retrieve roughly 50 prototype fixpoints per second.

\begin{table*}[t]
\caption{The outcomes of the benchmark. 
For each setting the time needed to fetch the fixpoints is shown in function of the number of fixpoints requested.
\emph{Single} is the benchmark with the server and the client on the same host and one simultaneous request. 
\emph{Multi} is the same settign with multiple requests.
\emph{Web Multi} is the setting where the client and server are on separate host with a simulated web link between them.
All timings are in milliseconds.}
\label{tab:remote} 
\centering
\setlength{\tabcolsep}{4pt}
\makebox[\textwidth][c]{ %
\begin{tabular}{rrrrrrrr}
\toprule
Amount & Single & Multi & Web Multi & Amount & Single & Multi & Web Multi  \\
\cmidrule(r){1-4}  \cmidrule(r){5-8} 
1,000 & 2,653  & 1,567   & 27,303  & 6,000 & 8,780  & 1,919  & 117,295  \\
2,000 & 2,889  & 1,190   & 37,715  & 7,000 & 10,181  & 2,124  & 131,140  \\
3,000 & 4,513  & 1,348   & 56,367  & 8,000 & 11,613  & 2,085  & 149,882  \\
4,000 & 6,042  & 1,381   & 74,976  & 9,000 & 13,493  & 2,227  & 168,415  \\
5,000 & 6,559  & 1,460   & 93,496  & 10,000 & 14,426  & 2,062  & 187,379  \\
\bottomrule 
\end{tabular} 
}
\end{table*}

\section{Conclusions}
In this paper we presented the software we developed to spearhead further work in the Prototype based knowledge representation.
We reviewed parts of the implementation of the prototype knowledge base and showed the results of benchmarks for consistency checking, fixpoint computing, and remote knowledge base access. 
At the current stage we see this software mainly used in research settings.
However, the benchmarks show that the software provides scalability needed for use in production environment.

\section*{Acknowledgments}
Stefan Decker would like to thank Pat Hayes, Eric Neumann, and Hong-Gee Kim for discussions about Prototypes and Knowledge Representation in general.

Michael Cochez performed parts of this research at the Industrial Ontologies Group of the University of Jyväskylä (Finland) and at the Insight Centre for Data Analytics in Galway, Ireland.

Furthermore, it has to be mentioned that the implementation of the software was greatly simplified by Google's Guava library, 
Apache HttpComponents,
the Jetty webserver,
Google GSON,
JUnit,
Apache Abdera, and
the Apache Commons Math™ library.

\bibliographystyle{splncs03}
\bibliography{sigproc,rfc}

\end{document}